\documentclass[conference]{IEEEtran}
\IEEEoverridecommandlockouts
\usepackage{comment}

\usepackage{cite}
\usepackage{amsmath,amssymb,amsfonts}
\usepackage{algorithmic}
\usepackage{graphicx}
\usepackage{textcomp}
\usepackage{xcolor}
\def\BibTeX{{\rm B\kern-.05em{\sc i\kern-.025em b}\kern-.08em
    T\kern-.1667em\lower.7ex\hbox{E}\kern-.125emX}}
\usepackage {subfigure}     
\usepackage{epstopdf}
\usepackage{subfloat}
\usepackage{color}
\usepackage[export]{adjustbox}
\usepackage{fancyhdr}

\renewcommand{\figurename}{Figure}
\begin{document}

\title{Impact of Feature Encoding on Malware Classification Explainability} 

\author{
	\IEEEauthorblockN{Elyes Manai}
	\IEEEauthorblockA{\textit{Department of Computer Science}\\ \textit {\& Software Engineering}\\
	\textit{Laval University, Quebec, Canada.}\\ Elyes.Manai.1@ulaval.ca}
	\and
	\IEEEauthorblockN{Mohamed Mejri}
	\IEEEauthorblockA{\textit{Department of Computer Science }\\ \textit {\&  Software Engineering} \\
	\textit{Laval University, Quebec, Canada.}\\ Mohamed.Mejri@ift.ulaval.ca}
		\and
	\IEEEauthorblockN{Jaouhar Fattahi}
	\IEEEauthorblockA{\textit{Department of Computer Science }\\ \textit {\&  Software Engineering} \\
	\textit{Laval University, Quebec, Canada.}\\ Jaouhar.Fattahi.1@ulaval.ca}
}

\IEEEoverridecommandlockouts

\maketitle
\IEEEpubidadjcol

\begin{abstract} 
This paper investigates the impact of feature encoding techniques on the explainability of XAI (Explainable Artificial Intelligence) algorithms. Using a malware classification dataset, we trained an XGBoost model and compared the performance of two feature encoding methods: Label Encoding (LE) and One Hot Encoding (OHE). Our findings reveal a marginal performance loss when using OHE instead of LE. However, the more detailed explanations provided by OHE compensated for this loss. We observed that OHE enables deeper exploration of details in both global and local contexts, facilitating more comprehensive answers. Additionally, we observed that using OHE resulted in smaller explanation files and reduced analysis time for human analysts. These findings emphasize the significance of considering feature encoding techniques in XAI research and suggest potential for further exploration by incorporating additional encoding methods and innovative visualization approaches.
\end{abstract}


\begin{IEEEkeywords}
Explainability, XAI, feature encoding, malware classification, preprocessing, LE, OHE, XGBoost. 
\end{IEEEkeywords}

\section{Introduction}
Machine learning has witnessed remarkable advancements in recent years, enabling the development of sophisticated models that achieve impressive performance on various tasks. As these tasks and the data they are trained on become more complex, so does the model complexity. This often causes the decision-making process to lack transparency, making it difficult to understand the reasons behind their predictions. In a society that uses AI for an ever-growing number of use cases, however, that lack of understanding can pose serious risks to the users. Averting these risks and allowing more control over what our AI is doing, thus allowing more responsible AIs, is the goal behind the Explainable Artificial Intelligence (XAI) subfield.  This subdomain of AI focuses on making black-box models transparent by providing understandable explanations for their decisions. XAI also allows us to combine the powerful pattern-recognition learning capabilities of AI with human-readable explanations that humans can instinctively understand and explain. the algorithms used in XAI usually work by finding out what parts of the input and of the model weights most affect the model's predictions. The end result will be a summary of each feature's contribution to the model. How helpful are these summaries, however, which we can call the quality of the generated explanations, depends on several parameters such as the chosen algorithm, the model architecture, and the data preprocessing technique. This last parameter, however, is not as popular as the others. While most XAI research focuses on algorithms, use cases, and the quality of explanations generated, there is a lack of research on the impact of preprocessing on generated explanations. We think that the preprocessing technique has a sizable impact on the quality of generated explanations and should be more explored. More specifically, we are interested in the feature encoding step of the preprocessing pipeline. Since XAI methods summarize feature contribution, the way we encode our models will directly affect the understandability of the generated explanations. Since preprocessing directly affects model performance, considerations must be taken to not trade off too much performance for better explanations, as better explanations on an unprecise model are not useful. Nonetheless, we think that a minor performance loss for a major boost in explainability is worth it, as it also opens up the door for better model and data understanding, bias discovery, robustness tests, and overall higher quality assurance. This is especially important in critical industries such as Medicine, Finance, and Cyber Security. To showcase the added value of our idea in a real use case, we will apply Machine Learning and Explainability on a common problem in Cyber security: Malware Classification. It is one of the most common tasks that Machine Learning is applied to in modern antiviruses and Intrusion Detection Systems. We will train a model on a publicly available malware dataset, apply the XAI algorithm, switch the preprocessing technique and compare the generated explanations. We will show that new rules and pain points can be detected and further explored by just changing the preprocessing technique. To the best of our knowledge, no prior studies specifically addressed the subject of the direct impact of preprocessing on explanation quality in the field of XAI have been identified in the existing literature. Our comprehensive review of the literature revealed that research in XAI is more geared towards XAI algorithms \cite{shap,lime,IG,anchors}, the generated explanations\cite{oneexplanation, quod}, alternative ways to bake explainability into the input features\cite{domain, ted} and other related problems\cite{focus, glocal}. Our focus in this paper can be summarized as follows: Given that XAI algorithms use the input features as the key components for the generated explanations, it is safe to assume that the type of feature encoding used will directly affect the clarity of the explanations. The more explicit the feature, the more detailed should be the explanation we get. With that in mind, we will study two main questions in this paper: 
\begin{enumerate}
	\item Does feature encoding affect explainability? 
	\item If yes, what encoding yields better explainability and why?
\end{enumerate}


\section{Concepts}

\subsection{Feature Encoding}
Feature encoding, also known as feature transformation or feature representation, is a crucial step in data preprocessing where categorical or textual features are converted into numerical representations that can be effectively used by machine learning algorithms. This is a mandatory step as ML algorithms only deal with numerical features. The choice of encoding technique directly impacts the ML performance. Here are some common feature encoding techniques:
\begin{itemize}
	\item \textbf{One-Hot Encoding:} Each category within a categorical feature is represented by a binary feature. If a feature has n categories, it is encoded into n binary features, where only one feature is active (1) for a particular category, and the rest are inactive (0). One-hot encoding is useful when there is no inherent order or relationship among the categories.
	
	\item \textbf{Label Encoding:} Label encoding assigns a unique numerical label to each category within a categorical feature. Each category is represented by a distinct integer value. Label encoding is suitable when the categories have an ordinal relationship or when using algorithms that can directly work with integer inputs.
	
	\item \textbf{Ordinal Encoding:} Similar to label encoding, ordinal encoding assigns numerical labels to categories. However, ordinal encoding takes into account the order or rank of the categories and assigns values accordingly. For example, "low," "medium," and "high" could be encoded as 1, 2, and 3, respectively.
	
	\item \textbf{Binary Encoding:} Binary encoding represents categories as binary bit patterns. Each category is assigned a unique binary code, and each bit in the code represents the presence or absence of a category. Binary encoding can be efficient for high-cardinality categorical features and reduces the dimensionality compared to one-hot encoding.
	
	\item \textbf{Embedding:} Embedding techniques are commonly used for encoding textual or high-dimensional categorical features. Embeddings are dense, low-dimensional representations that capture semantic relationships between categories. Embeddings are learned using techniques like Word2Vec \cite{DBLP:conf/somet/FattahiZM21, DBLP:conf/somet/FattahiMZG22} or categorical embedding layers in deep learning models \cite{DBLP:journals/access/DahoudaJ21}.
	
\end{itemize}

\subsection{Explainability}
Explainability in the context of machine learning\cite{10097489,9391727,ZiniA239667} refers to the ability to understand and interpret the decisions or predictions made by a machine learning model. It involves gaining insights into how and why a model arrives at a particular output, providing transparency and comprehensibility to the decision-making process. There are various approaches to achieving explainability:

\begin{itemize}
	\item \textbf{Model-Agnostic Approaches:} These methods aim to explain any black-box machine learning model without relying on its internal structure. They involve techniques like feature importance analysis, partial dependence plots\cite{pdp}, and surrogate models, which provide insights into the relationship between input features and model predictions.
	
	\item \textbf{Rule-Based Approaches:} These approaches aim to generate human-readable rules that describe the decision-making process of the model. Rule-based models, such as decision trees or rule lists, can provide explicit if-then statements that explain how specific features influence predictions.
	
	\item \textbf{Interpretable Model Architectures:} Some machine learning models, such as linear regression, logistic regression, or decision trees, inherently provide interpretable explanations. Their simplicity and transparency allow users to understand the impact of each feature on the final prediction.
	
	\item \textbf{Local Explanations:} Local explanation methods focus on explaining individual predictions rather than the model as a whole. Techniques like LIME\cite{lime} (Local Interpretable Model-Agnostic Explanations) or SHAP\cite{shap} (SHapley Additive exPlanations) provide insights into which features contributed the most to a particular prediction.
	
	\item \textbf{Visualizations:} Visualizations play a significant role in explaining complex models and high-dimensional data. Techniques like heatmaps, bar plots, scatter plots, or saliency maps help in visualizing feature importance, decision boundaries, or highlighting influential regions in the data.
\end{itemize}

\subsection{Malware Detection}
To demonstrate our work, we will take the common task of detecting malware. Malware are malicious pieces of software that are designed to infiltrate and damage information systems without the users' consent \cite{AslanS20, info11030168, Ameyed73, ZiadiaFMP20, Lou2020}. The term malware covers a lot of categories such as  viruses, ransomware, worms, trojans, backdoors, spyware, keyloggers, adware, bots, and rootkits. Malware analysts have to discover exactly what happened to a system and make sure that the machines damaged by malicious software are isolated from the organization's network. The analysis done to single out the suspicious parts of the software can sometimes take a group of analysts and several hours or even days. Since undetected malware can have devastating consequences on any organization, malware detection has been deemed one of the most important tasks in cybersecurity. Several types of systems have been built to detect and capture malware such as Intrusion detection systems, antiviruses and firewalls, and these systems keep getting smarter thanks to the combined shared knowledge of the cyber security community and the rapid advancement of technology. Current Malware detection systems use Machine Learning and Deep Learning to detect anomalies in files and network packets to protect the systems they're installed on. Since Machine learning has been known for its fantastic classification capabilities, more and more complex architectures and models are being tested and deployed to the current market. 
\section{Implementation and experimental results}

\subsection{Dataset}
For this project, we found a Malware classification dataset from the 2015 Microsoft Malware Classification Challenge\cite{microsoft_challenge}. The public variant we managed to download contains 19611 rows and 78 features. Each row represents a single file. The dataset is imbalanced as there are 14599 malware files and 5012 non-malware files, so 3 times as much malware.
The dataset has no missing data and all features are numerical aside from the "Name" one.

\subsection{Preprocessing}
The "Name" feature has been modified by the competition organizers to include "virus" if the file is malware and thus be removed since it does not represent real-life data. We do not apply any other preprocessing on the data aside from feature encoding. In this work, we apply two encoding techniques to all the features:
\begin{itemize}
	\item \textbf{Label Encoding:} Each feature value is represented by a unique integer.
	\item \textbf{One Hot Encoding:} Each feature value becomes a separate binary column where 1 means the file's value of that feature is the column name, and 0 if not. This allows for more precise knowledge of what went wrong.
\end{itemize}

\subsection{Machine Learning Modeling}
For training, we choose XGBoost\cite{DBLP:conf/somet/FattahiMZ21, DBLP:journals/cbm/LiuGG23} as our base model and train it using its default parameters, namely 100 estimators, a max depth of 5 and a learning rate of 0.1. We use the free Google Colab coding environment which offers a single sever with 12.7GB of RAM and a single NVIDIA T4 GPU with 15GB of GPU RAM.
To evaluate our model, We use four popular metrics: Accuracy, Precision, Recall and F1.
In a nutshell, accuracy measures the overall correctness of the model's predictions by calculating the proportion of correctly classified instances out of the total number of instances. Precision quantifies the proportion of true positive predictions out of all positive predictions made by the model, indicating the model's ability to correctly identify positive instances and minimize false positives. Recall measures the proportion of true positive predictions out of all actual positive instances in the dataset, representing the model's ability to capture positive instances and minimize false negatives. Finally, the F1 score combines precision and recall into a single metric by taking their harmonic mean, providing a balanced assessment of the model's accuracy and considering both false positives and false negatives. We showcase the performance results of XGBoost on the label encoded dataset in Table \ref{tab:xgb_results_full_data}.

\begin{table}[h]
	\centering
	\caption{XGB results on both encoding techniques}
	\label{tab:xgb_results_full_data}
	\scalebox{1}{%
		\begin{tabular}{|l|l|l|l|l|}
			\hline
			\textbf{Encoding Technique} & \textbf{F1} & \textbf{Accuracy} & \textbf{Precision} & \textbf{Recall} \\ \hline
			Label Encoding & 0.991 & 0.993 & 0.992 & 0.998 \\ \hline
			One Hot Encoding & 0.988 & 0.989 & 0.987 & 0.998 \\ \hline
	\end{tabular}}%
\end{table}

Although we did not preprocess our data, aside from encoding them differently, we managed to get pretty good results. We can therefore directly go to the explainability part.

\subsection{Explainability}
For starters, we are going to take away non useful features because one hot encoding all 77 features created 85102 features, which kept crashing our environment due to insufficient RAM. To do that, we will use XGBoost's built in feature importance function to list each feature's impact on the model's decision making. In Table \ref{tab:feature_importance_list}, we extract the top 10 influential features and sort them from most to least important.

\begin{table}[h]
	\centering
	\caption{Feature Importance according to XGBoost}
	\label{tab:feature_importance_list}
	\scalebox{1}{%
		\begin{tabular}{|c|c|c|}
			\hline
			\textbf{Rank} & \textbf{Feature} & \textbf{Importance score} \\ \hline
			1 & MajorSubsystemVersion & 0.6215 \\ \hline
			2 & Subsystem & 0.1362 \\ \hline
			3 & MinorOperatingSystemVersion & 0.0575 \\ \hline
			4 & MajorLinkerVersion & 0.0454 \\ \hline
			5 & SizeOfStackReserve & 0.0198 \\ \hline
			6 & Characteristics & 0.0100 \\ \hline
			7 & SectionMaxChar & 0.0084 \\ \hline
			8 & ImageBase & 0.0081 \\ \hline
			9 & SizeOfHeapReserve & 0.0081 \\ \hline
			10 & TimeDateStamp & 0.0059 \\ \hline
	\end{tabular}}%
\end{table}

According to Table \ref{tab:feature_importance_list}, the combined score of the 10 most important features are 0.9381 which means that they represent 93.81\% of the model's decision making power. We, therefore, can just keep these 10 features and not use the rest.
Doing so, we get the results shown in Table \ref{tab:xgb_results_top_10}.

\begin{table}[]
	\centering
	\caption{XGBoost results on the top 10 features}
	\label{tab:xgb_results_top_10}
	\scalebox{1}{%
		\begin{tabular}{|l|l|l|l|l|}
			\hline
			\textbf{Encoding Technique} & \textbf{F1} & \textbf{Accuracy} & \textbf{Precision} & \textbf{Recall} \\ \hline
			Label Encoding & 0.992 & 0.992 & 0.991 & 0.998 \\ \hline
			One Hot Encoding & 0.985 & 0.985 & 0.983 & 0.998 \\ \hline
	\end{tabular}}%
\end{table}

Comparing the results shown in Table \ref{tab:xgb_results_top_10} to those in Table \ref{tab:xgb_results_full_data} show that although we did lose a bit of performance, the drop is marginal (less than 1\%). This means that if the One Hot encoding does provide us with more explainability power, it would be recommended to use. 
For the next par, we will use a dedicated Explainability Algorithm called Shapley Additive Explanations (SHAP) to dig deeper into the model's inner reasoning. 

\subsubsection{The SHAP algorithm}
SHAP\cite{shap, SHAPAleneziL21} was introduced in 2017 and provides a unified way of explaining the contribution of each input feature to the final prediction of the model, based on calculated values called Shapley values. A Shapley value is a measure of the marginal contribution of a feature to the prediction, averaged over all possible combinations of features in the dataset. To calculate the Shapley values for a particular prediction, SHAP applies a game-theoretic approach based on the concept of cooperative games. It considers each feature value as a "player" in the game and computes the contribution of each player to the final prediction. It then calculates the average contribution of each player across all possible coalitions of players, weighting each coalition by its probability of occurrence. This approach results in a set of Shapley values, which represent the relative importance of each feature to the prediction for a specific instance. These Shapley values can be used to generate an explanation for the prediction, showing which features had the greatest impact and how they affected the final outcome. The mathematical formula used by SHAP to generate the Shapley Values is presented in Figure \ref{eq:shap_formula}.

\begin{equation}\label{eq:shap_formula}
	\phi_{i}(x) = \sum_{S \subseteq N \setminus \{i\}}\frac{{|S|!(|N|-|S|-1)!}}{{|N|!}}[f(x_S \cup \{x_i\}) - f(x_S)]
\end{equation}

Once generated, SHAP uses these values to display plots for both global explanations and local explanations.

\subsubsection{Global feature importance}
\begin{figure}[h]
	\centering
	\scalebox{0.42}[0.42]{\includegraphics{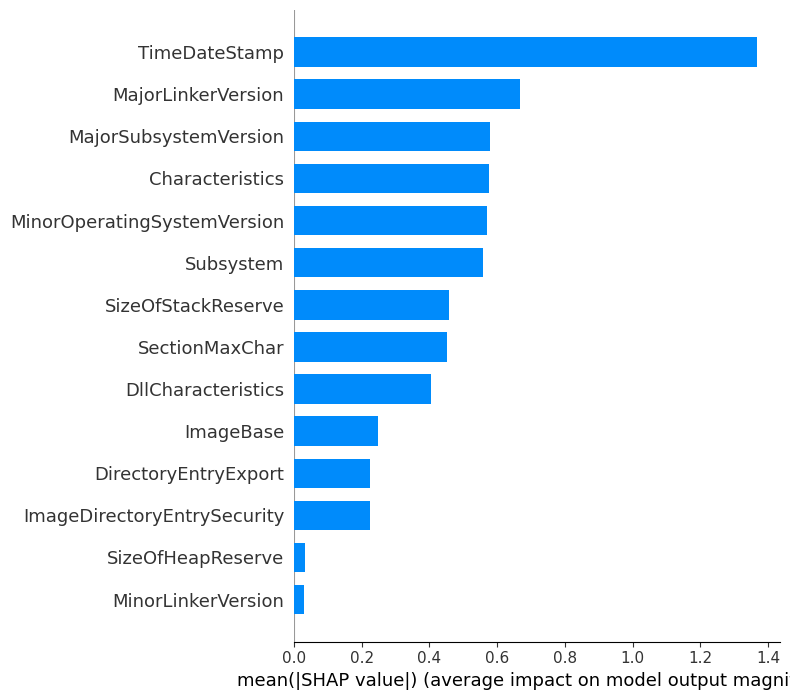}}
	\caption{Label Encoding global importance plot}
	\label{fig:label_shap_global}
\end{figure}

\begin{figure}[h]
	\centering
	\scalebox{0.42}[0.42]{\includegraphics{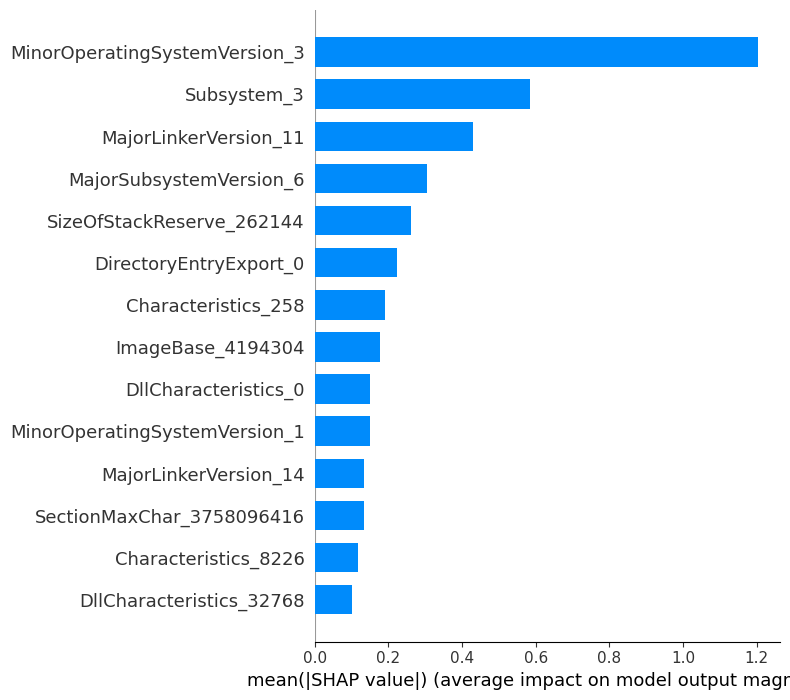}}
	\caption{One Hot Encoding global importance plot}
	\label{fig:ohe_shap_global}
\end{figure}

We use the SHAP algorithm to generate global summary plots that highlight the importance of each feature in the model’s decision-making similarly to what we have done in Table \ref{tab:feature_importance_list}. Figures \ref{fig:label_shap_global} and \ref{fig:ohe_shap_global} display the importance plots for the Label Encoded dataset and the One Hot Encoded dataset, respectively.

\section{Discussion}

\begin{figure}[h]
	\centering
	\scalebox{0.424}[0.5]{\includegraphics{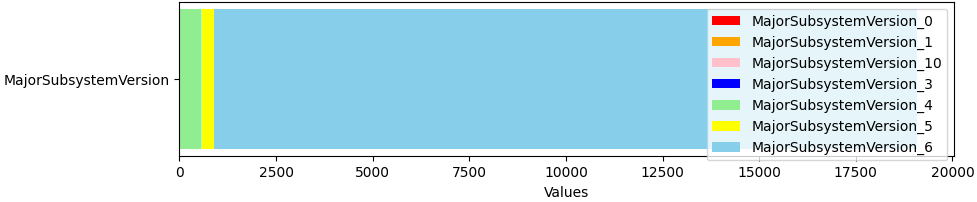}}
	\caption{Horizontally stacked bar plot of the "MajorSubsystemVersion"'s distinct values importance}
	\label{fig:horizontal_stack}
\end{figure}

\begin{figure}[h]
	\centering
	\scalebox{0.4}[0.4]{\includegraphics{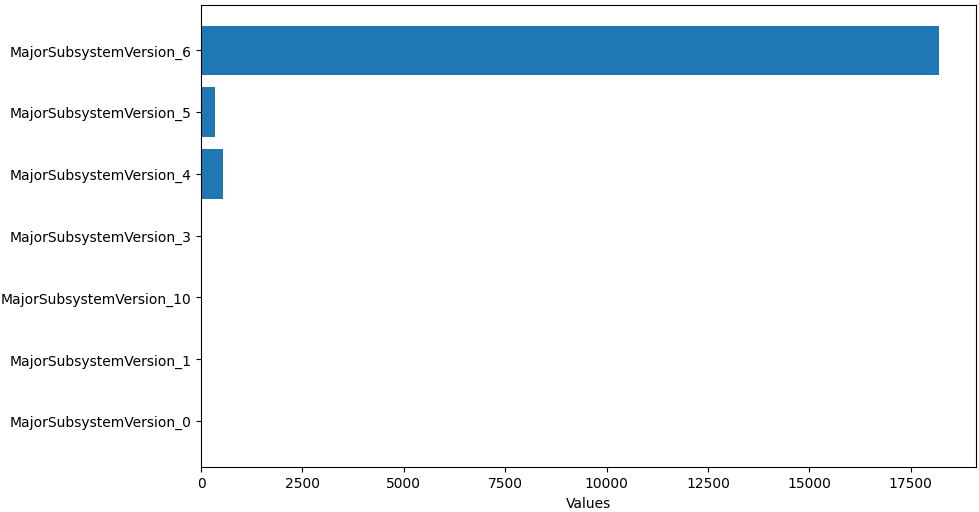}}
	\caption{Vertically stacked bar plot of the "MajorSubsystemVersion"'s distinct values importance}
	\label{fig:vertical_stack}
\end{figure}

The main difference between these plots is that while we know what feature is more important with Label Encoding, we know what exact value of that feature is more important with One Hot Encoding. This means that we get more specificity as a feature's importance is the sum of the importance of its unique values. A top ranking feature in the Label Encoding model could have therefore reached its rank because of the importance of some of its values, but not the others. Using One Hot Encoding, we can single out what values exactly are the most relevant to further analyze. For example, the "MinorOperatingSystemVersion" feature in \ref{fig:label_shap_global} has a mean SHAP value of almost 0.6, ranking fifth. However, in \ref{fig:ohe_shap_global}, we can see that is actually Version 3 of this feature that is really impactful, ranking first with a mean SHAP value of more than 1.2. Yet, version 1 of this feature only has a score of almost 0.2, and the rest of the version are not in the top 10 features. So using One Hot Encoding, we can single out files with the Version 3 of "MinorOperatingSystemVersion" and further analyze them separately in hopes of creating an easy rule for them or see what more we can learn. One drawback of this plot is that it is not easy to read when we have hundreds or thousands of features. In this example, we have 16087 features. It will be unproductive to use this plot to study feature importance. Instead, we can extract the raw SHAP values of all one hot encoded features, group them by original feature, and plot them side by side in another plot.  We propose the plots in Figures \ref{fig:horizontal_stack} and \ref{fig:vertical_stack} where we plot the importance of the different values of the "MajorSubsystemVersion" feature side by side, horizontally and vertically respectively. We chose this feature instead of the number 1 ranking "MinorOperatingSystemVersion" feature because it has considerably fewer distinct values making it easier to plot, wasting less space and delivering the same message. These figures allow us to better visually grasp the relativity in importance between the different values of a feature. This way, we can add or remove values to and from a watchlist and also construct rules for particular values. We can now combine this with the confidence score of the model at inference to start a routine, a check or apply a rule when the score doesn't hit the certainty thresshold. At that point, we would start investigating individual instances, thus needing different explanations called local explanations.

\subsubsection{local feature importance}
\begin{figure}[h]
	\centering
	\scalebox{0.45}[0.4]{\includegraphics{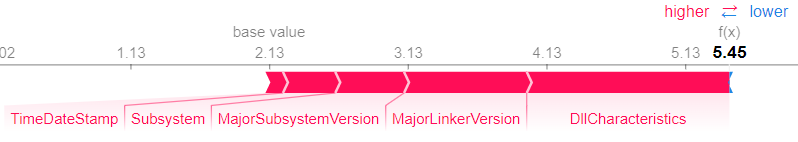}}
	\scalebox{0.35}[0.4]{\includegraphics{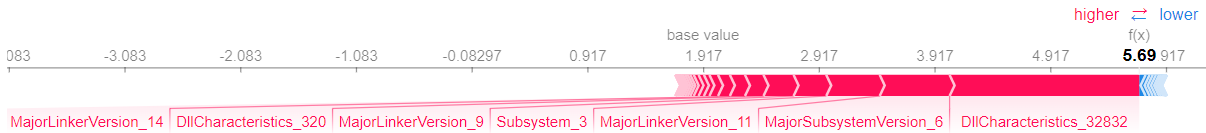}}
	\caption{Local explanation for test observation number 2}
	\label{fig:local_2}
\end{figure}

\begin{figure}[h]
	\centering
	\scalebox{0.35}[0.4]{\includegraphics{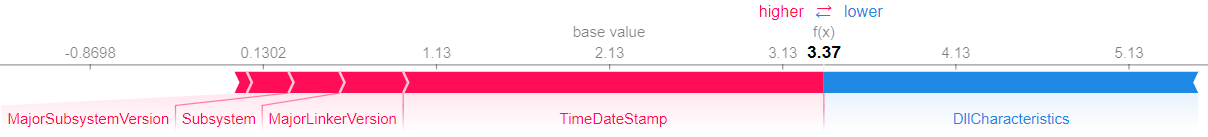}}
	\scalebox{0.325}[0.4]{\includegraphics{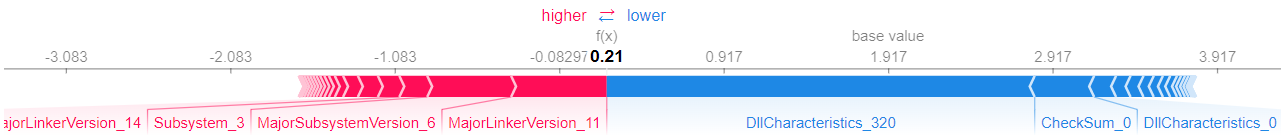}}
	\caption{Local explanation for test observation number 3}
	\label{fig:local_3}
\end{figure}

Local explanations focus on individual instances, displaying to the user the step-by-step contribution of each feature on the model's decision. Using SHAP's local explanation plots, we get Figures \ref{fig:local_2} and \ref{fig:local_3} which display the local explanation of instances 2 and 3 respectively, first using Label Encoding first and then One Hot Ecoding.

Again, the added refinement of the exact feature value gives us a lot more insight into what pushed the model towards a certain classification. Although the one hot encoding in this case may seem useless since we already know what value of each feature the instance holds, it instead can be used as an assertion method to make sure there are no anomalies in the decision shifting. Finally, we can see that being trained on the individual values changes the base value and decision shift intensity of each feature, as it has been trained on more finegrained data and the model had the chance to learn combinations that go together. These combinations in a tree based model such as XGBoost can then be used extracted and used as normal conditional IF rules or analyzed to detect vulnerabilities that went under the radar. Even then, the feature encoding will have an impact on the generated rules.

\subsubsection{IF-Rules}
\begin{figure}[h]
	\centering
	\scalebox{0.4}[0.45]{\includegraphics{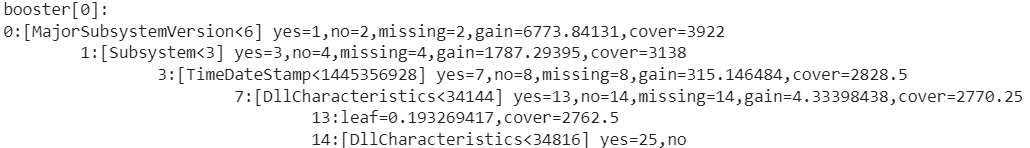}}
	\caption{Example IF-Rules for Label Encoding}
	\label{fig:rules_label}
\end{figure}

\begin{figure}[h]
	\centering
	\scalebox{0.4}[0.45]{\includegraphics{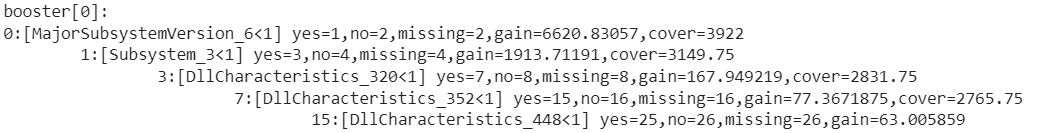}}
	\caption{Example IF-Rules for One Hot Encoding}
	\label{fig:rules_ohe}
\end{figure}

IF-Rules are logical statements that express conditional relationships between input variables and output decisions and follow a simple structure: IF a specific condition or set of conditions is satisfied, THEN a particular action or decision should
be taken. The conditions and actions are typically expressed using logical operators, such as "AND," "OR," and "NOT." IF rules provide a transparent and interpretable way to encode domain knowledge and decision-making criteria into a system. Due to their nature, tree-based models can be seen as a 
collection of IF rules combined together to form 
a decision-making process. Each node in a decision tree represents an IF statement on a specific feature or attribute, and the tree structure guides the flow of decision-making based on these conditions. The splitting criteria at each node determine the conditions for branching into different paths, leading to subsequent nodes or leaves with specific outcomes or predictions. Since XGBoost is a tree based model, we can extract the IF-Rules it learned during the training phase and use them to build logical pipelines or to study them. An example of the IF-Rules learned by our XGBoost model can be seen in 
Figures \ref{fig:rules_label} and \ref{fig:rules_ohe} for Label Encoding and One Hot Encoding respectively.

While there is no apparent difference between the IF-Rules of the two encoding techniques, 
the difference lies in the metadata. In Table \ref{tab:rules_diff}, we can see the difference in 
the rules' total text length in number of characters as well as the explanation file size in KB. We can see that One Hot Encoding resulted in less characters which means less file size.
The indirect consequence of this is less analysis time, less system complexity and less ambiguity, all of which directly benefit analysts and systems.

\begin{table}[h]
	\centering
	\caption{Difference in IF-Rules between the encoding techniques}
	\label{tab:rules_diff}
	\scalebox{0.9}{%
		\begin{tabular}{|l|l|l|l|}
			\hline
			\textbf{Encoding Technique} & \textbf{F1} & \textbf{Rules file size} & \textbf{Rules text length} \\ \hline
			Label Encoding & 0.991 & 221 KB & 226065 characters \\ \hline
			One Hot Encoding & 0.982 & 175 KB & 180237 characters \\ \hline
		\end{tabular}%
	}
\end{table}

\section{Conclusion}
In this paper, we studied the impact of feature encoding on the explainability of XAI algorithms. We took a malware classification dataset 
as an example on which we trained an XGBoost model. We tried two different types of feature encoding:
Label Encoding and One Hot Encoding and found there is a marginal performance loss
by the model by using OHE instead of LE. That loss was made up with thanks to the more detailed explanations
we managed to make thanks to OHE. We found that OHE allows us to go deeper in the details 
when searching for answers, both globally and locally. We also found that using OHE yields smaller explanation files and results in less time spent analyzing by human analysts.
We think this is an interesting aspect to be taken into consideration when working with XAI and could be 
expanded by including more feature encoding techniques and more creative plots.


\bibliographystyle{IEEEtran}
\bibliography{biblioart}

\section*{NOTICE}

\copyright 2023 IEEE. Personal use of this material is permitted. Permission from IEEE must be obtained for all other uses, in any current or future media, including reprinting/republishing  this material for advertising or promotional purposes, creating new collective works, for resale or redistribution to servers or lists, or reuse of any copyrighted component of this work in other works.
 
\end{document}